# What is a word?

Elliot Murphy[1,2]

1. Vivian L. Smith Department of Neurosurgery, McGovern Medical School, University of Texas Health Science Center at Houston, Houston, TX 77030
2. Texas Institute for Restorative Neurotechnologies, University of Texas Health Science Center at Houston, Houston, TX 77030

> *"Despite 2,400 years or so of trying, it is unclear that anyone has ever come up with an adequate definition of any word whatsoever, even the simplest."*
>
> (Elbourne 2011: 1)

In order to design strong paradigms for isolating lexical access and semantics, we need to know what a word is. Surprisingly few linguists and philosophers have a clear model of what a word is, even though words impact basically every aspect of human life. Researchers that regularly publish academic papers about language often rely on outdated, or inaccurate, assumptions about wordhood.

As in all scientific disciplines, we have two notions to consider:

1. Our *intuitive* concept of 'word' (which we all have, even though it can be vague, and sometimes hard to articulate fully, like most complex concepts).
2. A *technical, formal* definition of 'word' (which we can deploy to direct and inform experimental design, analysis and interpretation).

This is no different from other scientific concepts – for example, 'water' has a very intuitive meaning, but it also is linked to much more technical, formal notions emerging from chemistry and physics (Murphy 2023).

This short pedagogical document outlines what the lexicon is most certainly not (though is often mistakenly taken to be), what it might be (based on current good theories), and what some implications for experimental design are.

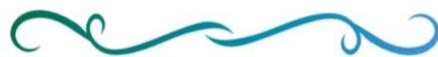

## What the lexicon is not

The central features of lexical items have no connection with sensorimotor instructions.

Lexical information is modality-independent, hence why we have sign language, braille, speech and writing. It's imaginable that, with some neural re-wiring, we could hook up some other modality to language, and comprehend language via olfaction, if our sense of smell was fine enough (not likely, but certainly possible) (Chomsky 2000).

Therefore, what we might call the 'phonological lexicon' and 'orthographic lexicon' are not in fact lexicons. They are sets of instructions for sensorimotor transformations – they get us *close* to lexical information, and are often the first steps in accessing it (Pustejovsky & Batiukova 2019; Woolnough et al. 2021).

It also therefore follows that distributional/statistical information about these ortho/phono 'lexicons' tells us something about how these modalities are organized – but it tells us nothing about how the lexicon itself is formatted. For example, if we know that 'the' is a high frequency word, and that 'broth' is a low frequency word, what does this tell us about their internal featural composition? Nothing at all.

Studies of grapheme-to-phoneme transitions are strictly focused on "performance systems"; or transformations of sensorimotor representations.

Another implication here is that the lexicon is not 'a dictionary in your head', 'a list of units in long-term memory', or some other such formulation. As we will see below, the lexicon is a *process*, not a thing. We do indeed have representations in long-term memory, but these look a bit different from what we typically think of as words. We have syntactic/categorial features in long-term memory, conceptual 'roots' in long-term memory, and we have 'form units' (a set of mappings from syntax to production-related representations) in long-term memory – all of these are separable, independent units, but they conspire to form what looks like on the surface something we usually refer to as 'the lexicon' (Adger 2022).

So what is left once we take away the orthographic lexicon and the phonological lexicon? How do we approach The Lexicon™ without resort to sensorimotor information?

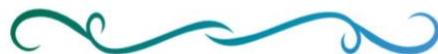

## What the lexicon might be

Lexical access immediately provides two sets of instructions: To sensorimotor systems (for externalization; articulatory phonetics, phonology, etc.) and to conceptual systems.

While the instructions to sensorimotor systems are intricate and complex, they are not necessarily domain-specific. For example, we can use any number of articulatory instructions, or execute any number of phonological computations (e.g., copy representation X and pronounce it alongside representation Y), without these being used for language. We can talk in gibberish, use a pencil to draw different types of shapes and intersecting lines – all without linguistic processes. There is still some debate in theoretical linguistics about whether phonology involves any domain-specific computations and representations, but most people assume that there is nothing intrinsically 'linguistic' about sounds or visual symbols (see *Glossa* Special Issue, 'Headedness in Phonology', ed. Shanti Ulfsbjorninn, 2017).

Another broader generalization from linguistics – that took thousands of years to prove and establish – is that the world's languages seem to differ *only* in their morphophonological features (Boeckx 2014). In other words, when we 'learn a second language', we only learn how morphology and phonology regulate the integration of meaning and structure (semantics & syntax). We do not 'learn' semantics or syntax. The basic semantic composition operations (e.g., *function application*, or M-join and D-join; Pietroski 2018), and syntactic combinatorial operations (e.g., *Merge, Agree*) are types of mental processes that are uniform across all languages – it is pretty much the definition of 'human' that you can execute these semantic & syntactic operations (Marcolli et al. 2023).

This conclusion allows us to make some predictions for how the instructions to conceptual systems will differ. One obvious prediction is that, unlike the instructions to sensorimotor systems, the instructions to conceptual systems *will* be characterized by certain processes and representations that are domain-specific, and language-specific, and difficult to capture with other systems (Dentella et al. 2023, Leivada et al. 2023). This prediction seems to be borne out, according to current thinking in generative linguistics. It also aligns with the idea that language is ultimately a system of *symbolic thought* (e.g., Chomsky, Dehaene, Hagoort, Tattersall, and many others).

The lexicon includes a number of features that facilitate the regulation of form/meaning mappings. These provide intricate instructions to various conceptual systems and 'core knowledge systems' (Spelke 2016), such as number sense, intuitive geometry, intuitive physics, theory of mind, etc. There are many such features, depending on the word in question (where 'word' should be taken to be 'a constellation of any such formal, semantic and syntactic features'). These include:

> **Argument structure features**: The number and type of logical arguments that a word can take, and how a word maps to syntactic expressions.
> **Event structure**: The type of event that a word can make reference to, e.g. a state, process, transition, whether the event is continuous or bounded, etc.
> **Lexical inheritance structure**: How a word semantically relates to other words. This is a core feature of more abstract words, that have no obvious sensorimotor grounding, and rely on their conceptual status mostly via associations with other clusters of semantic features.

Semantic features capture a number of roles that reflect human understanding of objects and relations (according to figures like Pustejovksy):

> **Formal**: The basic category that distinguishes an object within a larger domain.
> **Constitutive**: The relation between an object and its constituent part.
> **Telic**: The object's purpose and function.
> **Agentive**: Factors involved in the object's origin.

A simple example is the word 'book': The lexicon clearly does not contribute our sense of this word's color features, or shape features, or size features – these are all provided by non-linguistic conceptual and perceptual systems, and are built 'after' lexical access (during comprehension) once the brain has generated the basic structure of the semantic representation. What language *does* seem to uniquely encode is the Telic/functional information: the use and function of the object, the intention of its creation, means to a common end, and so on.

Language encodes things like 'evidentiality' (in Turkish, this is grammatically encoded), and other features relating to epistemology. Less emphasis is placed on things like color features (as far as I know, no language has morphemes for particular hues of certain colors, for example: these thoughts are encoded in full lexical items, not grammar). So different mental systems are 'hooked up' to language to varying degrees of emphasis. One can imagine an alien species who had our same language faculty, but one that was hooked up to different conceptual and perceptual domains. Their grammatical and syntactic information would be encoded differently, even if it came with the same limitations and constraints.

If we take away all of the non-linguistic features of words like 'book' or 'car' or 'blue' or 'run' or 'happy', what we are left with gives us an indication of what the lexicon contains – and what we find is this constellation of features that are present across various languages, such as semantic features and the word's syntactic category (Noun, Verb, etc.).

Along with these peculiar semantic features, we also have syntactic features, which are plainly language-specific (Adger & Svenonius 2012). We also have what is called a conceptual 'root' (Harley 2014): For example, the root RUN can be merged with an N feature to give us the noun 'run' ('That was a good run'), or it could be merged with a V feature to give us the verb 'run' ('I will run today'). Surprisingly, the majority of words in English are highly polysemous (Murphy 2021a, 2021b), such that their conceptual 'root' can be merged with different syntactic features. Many adjectives are good examples of this: 'red' is kind of adjective-like, but also kind of noun-like, depending on context. Philosophers have given around 80 or 90 different meaning for the word 'open': it is highly polysemous. The fact that we can run rampant like Shakespeare and make up our own verb-like use of 'red' (or any other word) suggests that the lexicon regulates the mapping of basic conceptual roots with an associated list of semantic and syntactic features (see *The Linguistic Review* Special Issue, 'Roots in Context', ed. Noam Faust and Andrew Nevins, 2019).

For example, 'destroy' (DESTROY + V) can be further categorized via nominalization, turning it into a noun (destruction = [DESTORY + V] + N]). This involves a root being merged with a V feature, and then the 'lexical item' being again merged with an N. This has often been how the English language has evolved over the centuries: verbs get nominalized, and nouns also get verbalized. But we don't have two separate lexical items resulting from this, we just have one compositional morphological structure, a 'word'.

From an evolutionary point of view, this also makes rough sense: non-human primate cognition is conceptually rich, and human evolution seems to be defined by the mapping of this pre-linguistic primate conceptual system (where we get some 'roots' from) to some strange new system of categorization and hierarchy, putting together roots with some identity like N or V (Murphy 2019).

These basic conceptual representations/roots need to be paired with some syntactic feature, and then trigger various semantic instructions, in order to count as a 'word'. A good model of the lexicon is that it pairs conceptual representations with semantic and syntactic features, then it sends these off to sensorimotor systems (orthography, phonology, etc).

The order of operations is something like this:

1. Select root.
2. Merge root with syntactic features (Noun, Verb, Preposition, Determiner, Adjective, Tense..., but also phi-features like Person, Number, Gender, and Case features, depending on the representation being fetched).
3. Send to conceptual systems and sensorimotor systems.
4. Conceptual systems assemble appropriate features (Telic features, event features, animacy, theta-role features like Agent, Patient, Instrument, etc).
5. Sensorimotor systems (i.e., the 'orthographic lexicon' and 'phonological lexicon') provide their own set of modality-specific features to externalization.
6. Word is interpreted and pronounced.

What, then, is a 'lemma' meant to be? It is assumed to be the *first step* in the interface of the lexical selection process and the mapping to sensorimotor systems. The most basic and regularized form that needs to be mapped to some meaning. You will see below why the idea of the lemma needs to be challenged, but for now we can try and map some more traditional notions onto this framework we're developing here.

'Run' can never be accessed in a superposition: it is always accessed as either a verb or noun. We can never access the conceptual root and articulate it without first pairing it with a syntactic/categorial feature.

During this process, there are basically two content types: nouns and verbs. Adjectives are kind of like nouns, and adverbs are kind of like verbs, they just modify properties of the object or event. Notice that these are derived from the basic conceptual structure of the human mind – events and objects, the fundamental conceptual divide. This is a very simple observation, but it helps reinforce the basic architecture here: the lexicon is in the game of directly mapping basic conceptual roots to a format that is interpretable and usable by other mental systems.

We also have *functional grammatical structure* (function words) that encode how N and V can relate to each other, like determiners and prepositions and complementizers. These might look very different, but they are treated the same during lexical access: we pair a root with some syntactic features, but *also* semantic features. Function words still have a meaning, after all. They don't provide rich instructions to variable conceptual systems, unlike words like 'lunch' and 'nostalgia', but they do provide some fairly rigid logico-semantic instructions, though ones which are difficult to formalize: Barry Schein (2017) wrote 1040 pages just on the word 'and'.

Consider idioms as another example here. These show how the regulation of form/meaning mappings cuts across both content and function words. We can have 'kick the bucket' to mean 'die', but we can also have structures like 'they are to remain here' and 'am I to go there alone?', which involve functional elements triggering functional semantics like Tense ('finite' feature) being merged with a root 'BE' to yield a functional, logical meaning.

The process of lexicalizing a concept is an intricate one. It does not simply provide a 'form' to a concept (pronounce concept X using phonological representations A, B, C). It also seems to provide a new kind of format to the concept, to imbue it with certain features that the concept otherwise wouldn't have (like Pustejovsky's features above, but also some other peculiar ones). Lexicalizing a concept *changes it*.

To recap some of this: Once a concept is lexicalized for the first time it can be part of the "semantic lexicon" (the roots are being mapped to some syntax, and then to a form). But we also have long term storage of syntactic and semantic features, which are paired with concepts to form feature bundles. These bundles of features then provide instructions to performance systems in the form of orthography and phonology. But notice that what we usually call "a lexical item" is actually the read-out of this process of merging conceptual roots with categorical and formal features.

Also, not all potential roots are going to be lexicalized. The human mind likely stores plenty of conceptual roots that are yet to be accessed appropriately by the language system – and indeed, many languages express 'thoughts' that are not directly lexicalized in English, often relating to more nuanced thoughts like subtle emotions, experiences, and interpersonal relations. ("In my language, we have a word for people like you...", etc).

Returning to the lemma model issue: 'Run' is the lemma of runs, run, running. But 'lemma' is a purely morphological concept, pertaining to externalization systems. It interfaces directly with lexicality and syntax-semantics, because language is about the regulation of form and meaning. Establishing form/meaning pairs is what the process of lexical access is all about.

Things get more complex when we take a root like 'go', and merge it with a V feature and a Tense feature (Past), to generate the form 'went' – which looks nothing like 'go', but is the lexical transformation we get based on the specific syntactic and semantic features we compose the root with. The 'elsewhere condition' governs a lot of these cases: we have marked and unmarked forms (default forms), with some deviations. Markedness as a formal concept does not just apply to sound systems, as was thought in the early 20th century; semantic representations also have their own markedness criteria (Murphy 2023).

The lemma is just the base form-meaning pair that is then subject to further morphological transformations to drive various lexemes.

Lexicality is therefore about syntax-morphophonology mapping.

Syntactic features form the *basis* of mapping to morphophonology and the *basis* of mapping to semantics, but this is far from transparent. Language does not actually pair individual forms with individual meanings, nor does it pair individual syntactic structure with individual meaning (polysemy).

The process is something like this:

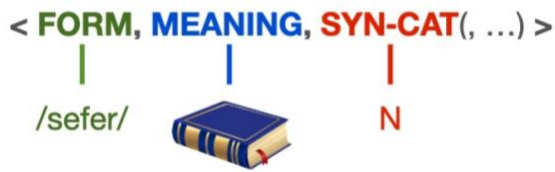

But this is not quite right, for reasons mentioned above: the process of the lexicon does not pair individual forms with individual meanings. The system is much more dynamic, being able to pair form X with meanings X, Y and Z (polysemy). At the same time, language *also* does not pair syntactic structure X with meaning A, since we have well-known cases of syntactic ambiguity ('I watched a movie with Jim Carrey' can mean we watched a movie starring Jim, or we watched *Syndromes and a Century* whilst sitting next to Jim).

This tells us something quite deep: the mapping of lexical items and assemblies of lexical items ('sentences') to meanings is non-trivial, and certainly not one-to-one. So, language seems to provide instructions to conceptual systems that can be used in multiple ways, depending on context or constraints imposed by other mental faculties. The language system provides different perspectives for interpreting the mind-independent world, and affords new directions for planning, inference and subjective reflection (Leivada & Murphy 2021).

To conclude: 'lexical access' is a process of pairing a root with a syntactic/categorial feature (N, V, A, P...), additional syntactic features (Tense, phi-features...), and then sending this structure to sensorimotor systems for morphophonology (accessing orthographic and phonological lexicons, etc) and conceptual systems for interpretation and the assignment of meaning (theta-roles, types of conceptual interpretation).

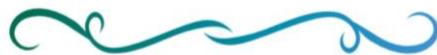

## What the lexicon looks like to the experimentalist

In the classical architecture of language production, when we see a picture of a car and then ultimately say the word 'car', the first stage of Conceptualization fetches all the non-linguistic conceptual information that we briefly mentioned above (shape features, color features, size features, etc.) in order to trigger the accessing of a conceptual root, CAR. Then, the process of Lexical Access involves the pairing/merging of CAR with syntactic/categorial features, like N (CAR + N), and any other syntactic features (phi-features, like Number). A word does not become a word until we have paired a root with some category (ALL words have a category – they are not words if they are category-less) and some semantic or formal features.

Only then can we proceed to Formulation. Notice that under the framework I am proposing here, Formulation is strictly at the sensorimotor systems, so it cares about morphophonology, or how the instructions provided by the lexical access process (CAR + N + Number features, etc.) can be successfully passed over to sensorimotor transformations.

Each of these lexical features are interpreted by either sensorimotor or conceptual systems. For example, 'Plural' has a conceptual interpretation but it also obviously provides a clear morphophonological instruction (usually, add 's' to the end of the form, like 'cars').

We end up here with no traditional notion of 'word', no clear boundary between classical linguistic structures, no notion of one-to-one form-meaning mappings. All we have is a process of pairing roots with syntactic and semantic features, and then mapping each of these features to some instruction at either the phonological/orthographic lexicons, or the conceptual system, or both.

For example, in English we can say a sentence like 'It is completely dry'. 'It' has a syntactic feature and a phonology, but is light on semantic features. It has no conceptual semantic instructions, it just pinpoints some entity. So this is a case of [+syntax, +phonology, -semantics]. In Italian, we can simply say 'E completamente secco', which translates as 'is completely dry'. We don't need to say 'it'. Other words have [-syntax, +phonology, +semantics], like 'Hello' and 'Ouch!', with are simple statements with no obvious syntactic feature. Lastly – and most interestingly of all – we have elements that are [+syntax, -phonology, +semantics], or units of linguistic computation that we don't pronounce overtly but we used to construct syntactic and semantic information, like in 'Mary went to the park and John did [ ] too', where the [ ] element is copied from previous discourse.

There is no clear boundary between words and phrases and structures: many words are kind of like 'mini sentences', like anti-institutionalization, which is extremely hierarchically complex, and involves just as many instances of merging units together as a sentence like 'the red boat sank'. Yet, we can call it a 'word' because we can top the whole derivational process off with a final [+N] merging operation, which categorizes the whole unit, whereas in 'the red boat sank' we don't do this – unless a movie is made called *The Red Boat Sank*, in which case we can apply an [+N] lexicalization process, and store the unit in memory, and use it just as we would any other noun ('John likes it'; 'John likes *The Red Boat Sank*').

Morphology and syntax are no different, under this perspective. There is only 'morphosyntax'. Putting features together into hierarchical, labeled units, leading to compositionality all the way down (Murphy & Shim 2020) – the whole is a function of the parts and the way in which they are combined, not just at the sentence level, but also at the 'word' level. For example, within polysynthetic languages (like Inuktitut, an Eskimoan language) there is of evidence for non-linear (hierarchical) relations between the elements within morphologically complex words. Morphemes merged together are not simply beads on a string, a linear concatenation of elements: they form relations between each other based on hierarchy. For example, in Inuktitut we can use 'havautituqtiniaqtara' as a single morphologically complex word to mean a very complex meaning ("I'm going to give her medicine frequently"), whereas in English we have to say multiple words (we could of course create a new word for this meaning, too). But this leads to a problem for semantics: a single word isn't supposed to be able to generate a compositional meaning, referring to some complex event structure with agents and patients, etc. So we should abandon all hope in a simple lemma model of lexical access (Krauska & Lau 2023).

Some problems arise for models pretty immediately: for example, Matchin and Hickok (2020) assume that words are 'treelets', with their morphemes being composed hierarchically. However, this model still assumes that 'words' themselves have some kind of computational status that is independent of other levels of linguistic structure. The brain might not ultimately care if something is a 'word' or a 'morpheme' or a 'sentence' – it just wants to combine features pertaining to form, meaning and syntax, with many possible combinations arising here (Murphy et al. 2022, 2023), although at the same time the undeniable impact of lexical frequency on cortical response profiles (Woolnough et al. 2021) also forces us to acknowledge the quasi-'lemma-like' formatting of online lexical processing (see next section).

One of the more obvious implications for experimental design that emerges here is the need to consider relevant semantic and syntactic features during 'lexical access', alongside the more common distributional statistical measures like neighborhood measures and frequency, etc.

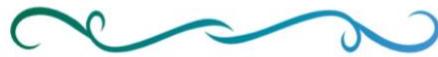

## Krauska & Lau (2023): A brief case study

An excellent resource to consult in this connection is Krauska & Lau (2023). A nice passage from this paper is copied below (the term 'non-lexicalist' that is used here is commonly deployed in the linguistics literature to refer to any model that does not assume that the lexicon is a stored list of words with one-to-one mappings between form, syntax and meaning; for example, the type of model I outlined above is clearly non-lexicalist):

> "Instead of relying on a lemma representation, a non-lexicalist production model can represent stored linguistic knowledge as separate mappings between meaning and syntax, and syntax and form, such that meaning, syntax, and form may not line up with each other in a 1-to-1-to-1 fashion. Such a model can also account for prosodic computations that depend on meaning, syntax, and form information. Furthermore, we suggest that cognitive control mechanisms play an important role in resolving competition between the multiple information sources that influence the linearization of speech."

These authors also review some problems with the lemma model:

> "In polysynthetic languages, a single word can be composed of many productive morphemes, representing complex meanings. In order to represent those words as lemmas, each lemma would have to correspond to very complex lexical concepts, with many redundant lemmas, to represent all of the possible morpheme combinations in that language; alternately, each lemma would have to incorporate a massive set of features in order to have a "complete" inflectional paradigm."

> "The evidence from Inuktitut and Vietnamese indicates that, not only do we need to move away from a view of production in which stored lemmas correspond to words, but we also need to give up the idea that the units of language production are syntactically atomic by definition."

> "There is a large amount of data that the lemma will struggle to model, including (but not limited to) inflection and morphological structure, suppletion, and idioms, phenomena which are fairly widespread throughout human languages. These phenomena suggest that syntax and morphology need to be able to interact fully, not just by sharing a limited set of features, and that the form and meaning of a syntactic object is partially determined by the syntactic context, not just by the syntactic object itself."

An important caveat: 'lemmas' might indeed exist in some languages, in some cases, where there is a transparent and unambiguous mapping between form, syntax and semantics, but if these representations do arise, it is kind of by accident, not because 'lemmas' are a core component of the psycholinguistic basis of language.

> "So far in this section, we have argued against the claim that the system of language production requires lexical knowledge to be formatted in terms of lemmas or lexical units as an organizing principle. However, for things that do have a 1-to-1-to-1 mapping between meaning, syntax, and form (where a single syntactic object has a consistent meaning and form across a variety of contexts), it would be entirely plausible that lemmas—or something like them—could arise as a byproduct of language-specific optimization, where it would be faster or more efficient to represent meaning, syntax, and form in that way, even if it is not an architectural principle. In these cases, it is possible that the translations which are performed for that word can treat the word as if it were atomic (i.e., the calculation to determine the form for the word does not need to refer to any other elements in the syntactic context), as is suggested by the lemma model. This kind of symmetry might occur more often in some languages, so linguistic behavior may appear to be more "lemma-like" than it would for other languages. To be clear, this would be a consequence of optimization at the implementation level, rather than the representation or algorithm level."

> "Depending on the properties of a particular language, storage of different sized pieces may optimize production, allowing wide variation cross-linguistically in the size of the stored pieces even if the underlying grammatical architecture is assumed to be the same."

Krauska & Lau (2023) provide a neat model of language production that is much more cognitively plausible and in line with current thinking in linguistics (for discussion, see also Murphy 2024).

> "We assume instead that linguistic knowledge includes sets of syntactic atoms, sets of mapping rules between syntactic units and meaning units, and sets of mapping rules between syntactic units and form units (Preminger, 2021). The syntactic terminals are fully abstract, meaning that they have no form or meaning themselves; both their meaning and form are conditioned by their context within the syntactic structure. The two sets of mappings may not necessarily be "symmetrical," in that for a single component of meaning which maps to a piece of syntax (however complex), that piece of syntax may not map to a single form segment; conversely, for a single form segment which maps to a piece of syntax, it may not correspond to a single component of meaning […] Furthermore, it is also possible in this model for a piece of syntax to have no mapping to meaning (for example, the expletive it in a sentence like "it is raining" has no possible referent) or no mapping to form (such as phonologically null elements)."

"This model is non-lexicalist because the mechanisms which generate the syntactic structure make no distinctions between processes that apply above or below the word level, and there is no point at which meaning, syntax, and form are stored together as a single atomic representation. Each stage in the model is a translation between different kinds of data structures. The "message" integrates different components of non-linguistic cognition, including memory, sensory information, social dynamics and discourse information, theory of mind, and information structure. Translating that message into a syntactic structure means re-encoding the information into a format that is specifically linguistic, involving syntactic properties and relations that may not be transparently related to the intended message. The hierarchical structure of syntax, in turn, must be translated into a series of temporally ordered articulatory gestures in order to be uttered as spoken or signed language."

This model is basically a rethinking of the Conceptualization, Lexical Access and Formulation stages:

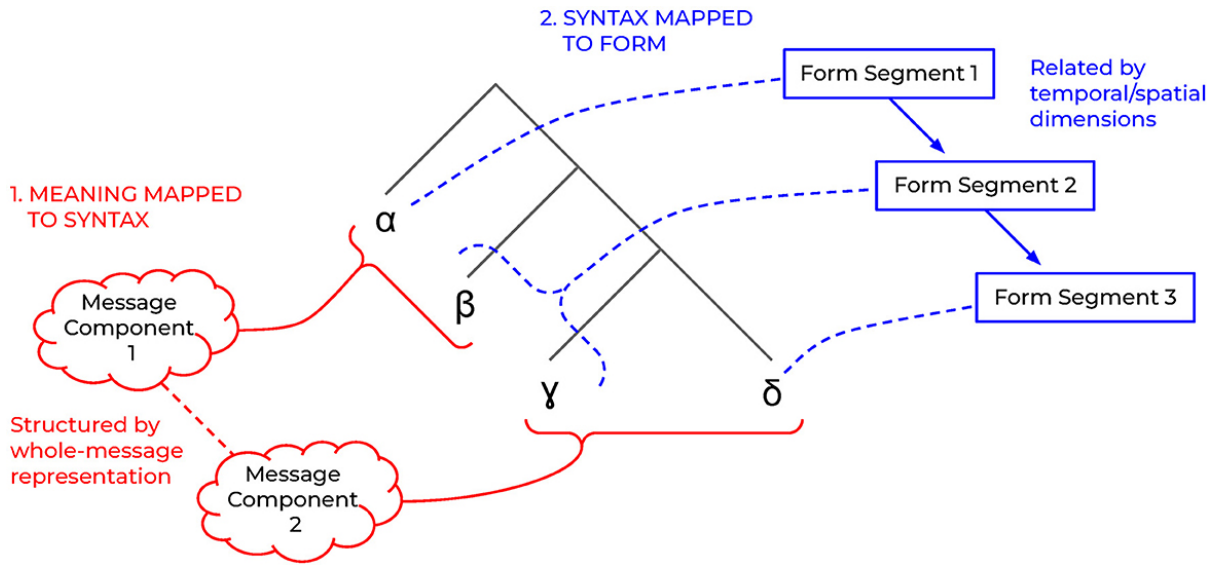

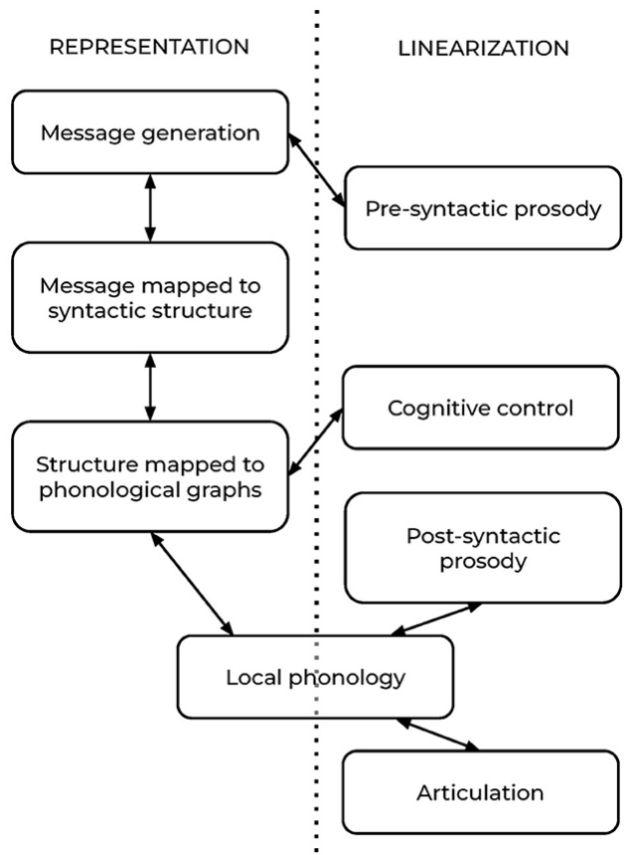

The stages of language production for Krauska & Lau (2023) are:

1. Message generation
2. Message mapped to syntactic structure
3. Pre-syntactic prosody
4. Syntactic structure mapped to segments of phonology
5. Cognitive control
6. Local phonology and phonological buffer
7. Post-syntactic prosody
8. Articulation

Only (2) here is truly language-specific. (4) also involves some kind of linearization algorithm, to compress hierarchical structures into some flat object for externalization. All the other processes call upon various brain systems (Murphy 2015, 2020, 2024). Every process here is roughly 'linguistic' in the broad sense, given the context of language production, but experiments that isolate (2) will evince neural processes that are likely to be more specific to language.

Other reflections are given on aphasias, under this model: A deficit in syntax does not necessitate deficits at any other levels (concepts, or form). And same for other things: a deficit in accessing form does not predict deficits in concepts or syntax:

> "As long as a given string is phonologically well-formed, as external observers we may not necessarily know if it was also syntactically well-formed. By moving away from the "triad," just knowing that a phonological word was correctly produced may not be indicative that its meaning and syntax were also correctly generated, only that a form was produced. The only part we have direct access to is the utterance. For that reason, testing theories of aphasia may require more careful thought about what other processes may be at work beyond the one mechanism which is impaired, and how they might hide the real deficits."

Lastly, our intuitive notion of wordhood can often be derived from sensorimotor biases:

> "It could be that most of our intuitions about wordhood are in fact grounded not in natural spoken language, but in orthography, among literate communities whose writing system make use of white spaces as separators. For readers of such orthographies, "word" could serve as a useful term for the things between white spaces, which might well define processing units for the reading modality. However, many other writing systems have not made use of this convention, and it is notable that those speakers often have much less developed intuitions about wordhood (Hoosain, 1992). In summary, it is hard to see how speakers' intuitions about wordhood systematically correspond to any representational or processing unit of natural spoken language, although they could correspond to units of certain written languages."

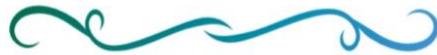

## Conclusion

Linguists no longer assume that words are the basic units of linguistic structure-building. Instead, it is groups of features, some of which look like what we intuitively think of as words, but there are others which look totally different.

Syntactic and morphological processes are not distinct – there is no way to draw a line between them, other than via notational conventions and/or formal apparatus. The language system simply fetches roots, then syntactic/categorial features, other relevant features, and ships them off to two different interfaces for externalization (phonology, orthography, etc.) and internalization (interpretation, conceptual information, meaning, etc.).

Language is a system of recursively combining features into hierarchies. There is no need for us to keep strictly to classical Aristotelian and intuitive notions like words and phrases – these are often useful for informal description, but not theoretical explanation. These feature combinations provide instructions to two systems, for internalization and externalization. These systems then have their own domain-specific constraints and biases and limitations (much of linguistics is precisely about what these semantic and phonological constraints are), but language itself is at its core a system of combinatorics and hierarchy. The primary focus here is therefore syntax-semantics, and not anything else (e.g., types of 'form').